%% file: egpaper_final.tex
\pdfoutput=1
\documentclass[10pt,twocolumn,letterpaper]{article}
\usepackage{float}
\usepackage{iccv}
\usepackage{times}
\usepackage{epsfig}
\usepackage{booktabs}
\usepackage{graphicx, multicol, subcaption}
\usepackage{amsmath}
\usepackage{amssymb}
\usepackage{authblk}
\usepackage[accsupp]{axessibility}  
\graphicspath{{images/}}
\input{macrosjvg}

\usepackage[breaklinks=true,bookmarks=false]{hyperref}

\iccvfinalcopy 

\def\iccvPaperID{3} 

\ificcvfinal\fi

\begin{document}

\title{Investigating transformers in the \\ decomposition of polygonal shapes as point collections}

\author[]{Andrea Alfieri}
\author[]{Yancong Lin}
\author[]{Jan C. van Gemert}

\affil[]{Computer Vision Lab,
Delft University of Technology, the Netherlands}

\maketitle
\ificcvfinal\thispagestyle{empty}\fi

\begin{abstract}
   Transformers can generate predictions in two approaches: 1. auto-regressively by conditioning each sequence element on the previous ones, or 2. directly produce an output sequences in parallel. While research has mostly explored upon this difference on sequential tasks in NLP, we study the difference between auto-regressive and parallel prediction on visual set prediction tasks, and in particular on polygonal shapes in images because polygons are representative of numerous types of objects, such as buildings or obstacles for aerial vehicles. This is challenging for deep learning architectures as a polygon can consist of a varying carnality of points. We provide evidence on the importance of natural orders for Transformers, and show the benefit of decomposing complex polygons into collections of points in an auto-regressive manner.
\end{abstract}




\input{figures/datasets}

\input{sections/0_introduction}

\input{sections/1_related_work}

\input{figures/model}

\input{figures/embeddings}
\input{sections/2_method}

\input{sections/3_experiments}

\input{sections/4_conclusion}

{\small
\bibliographystyle{ieee_fullname}
\bibliography{egpaper_final}
}

\end{document}

%% file: macrosjvg.tex
\usepackage{comment} 
\usepackage{xcolor}

\newcommand{\Eq}[1]{Eq.~(\ref{eq:#1})}
\newcommand{\eq}[1]{\Eq{#1}}
\newcommand{\fig}[1]{Fig.~\ref{fig:#1}}
\newcommand{\tab}[1]{Tab.~\ref{tab:#1}}

\usepackage{graphicx}
\usepackage{amsmath}
\usepackage{amssymb}
\usepackage{booktabs}
\usepackage{comment}


%% file: figures/datasets.tex
\begin{figure*}[t]
    \centering
    \begin{subfigure}[b]{\textwidth}
        \centering
        \begin{minipage}{0.237\linewidth}
        \includegraphics[width=0.99\linewidth]{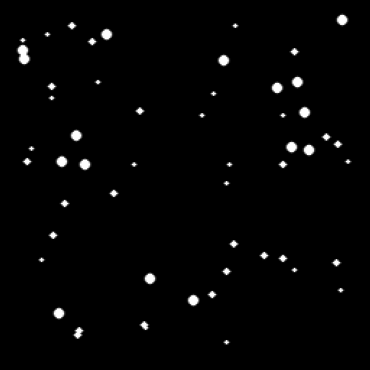}
        \captionof{figure}{Points}
        \end{minipage}
        \hfill
        \begin{minipage}{0.237\linewidth}
        \includegraphics[width=0.99\linewidth]{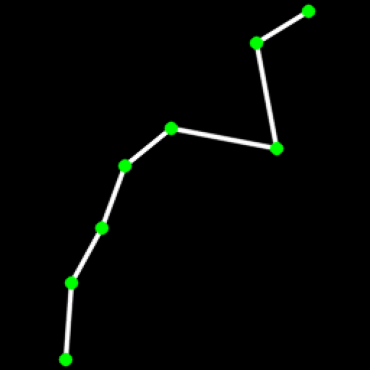}
        \captionof{figure}{Lines}
        \end{minipage}
        \hfill
        \begin{minipage}{0.237\linewidth}
        \includegraphics[width=0.99\linewidth]{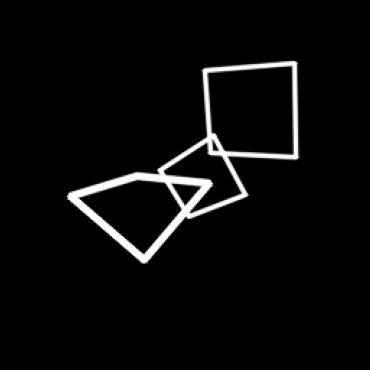}
        \captionof{figure}{Gates}
        \end{minipage}
        \hfill
        \begin{minipage}{0.237\linewidth}
        \includegraphics[width=0.99\linewidth]{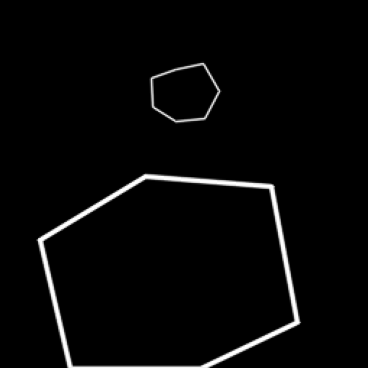}
        \captionof{figure}{Polygons}
        \end{minipage}
    \end{subfigure}
    \vskip\baselineskip
    \begin{subfigure}[b]{\textwidth}
        \centering
        \includegraphics[width=0.237\linewidth]{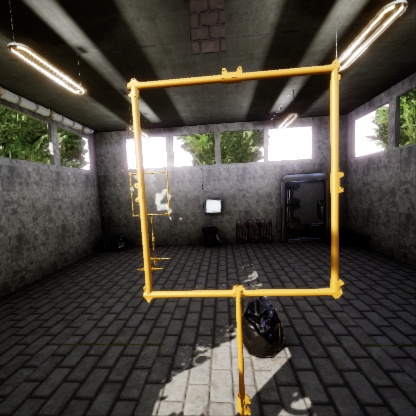}%
        \hfill
        \includegraphics[width=0.237\linewidth]{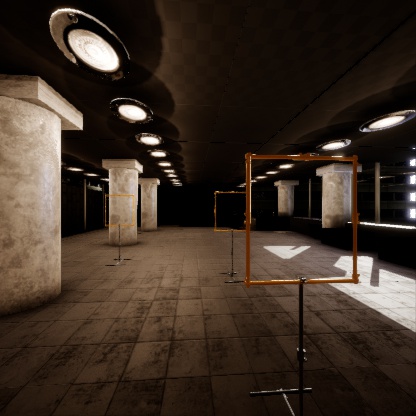}%
        \hfill
        \includegraphics[width=0.237\linewidth]{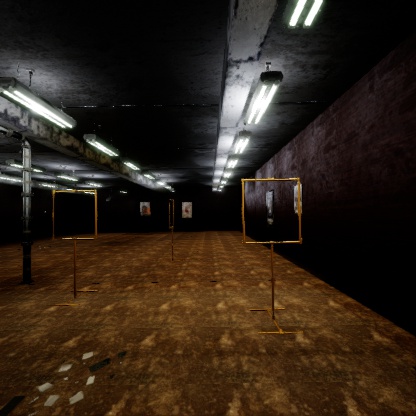}%
        \hfill
        \includegraphics[width=0.237\linewidth]{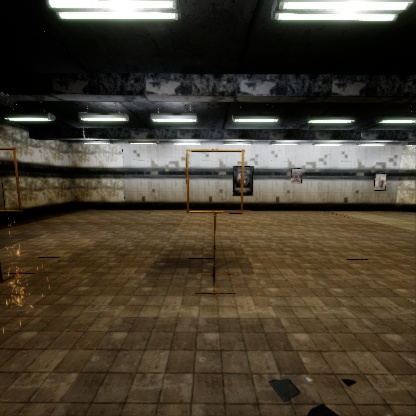}%
        \caption{Synthetic gate dataset. }
    \end{subfigure}
    \caption{\textbf{Examples of point collections used in our paper.}}
    \label{fig:datasets_examples}
\end{figure*}

%% file: sections/0_introduction.tex
\section{Introduction}
Predicting polygonal shapes in images is a high-level task that is representative of many vision problems. One example is to automatically extract vectorized building outlines from high-resolutions satellite images \cite{girard2020polygonal, girard2018end, iglovikov2018ternausnetv2, li2019topological}. Similarly, polygon detection can be beneficial for vision-based flight control of unmanned aerial vehicles (UAVs), as demonstrated in the Autonomous Drone Racing \cite{jung2018direct}, where drones fly autonomously through a sequence of polygon-shaped gates purely relying on vision signals. The performance of polygon detection is critical for UAVs as it directly affects the precision of navigation. 

In this work, we treat polygonal shape prediction as a \emph{collection prediction} task such as sequences and sets . Many kinds of data can be naturally represented using collection and many machine learning tasks can be viewed as a collection prediction problem, such as predicting the collection of points of a polygon \cite{zhang2019fspool}, detecting objects in an image \cite{carion2020end, zhang2019deep}, estimating the pose of humans by detecting a set of key-points \cite{cao2019improving, tang2019does}, or predicting multiple labels for the same sample \cite{rezatofighi2017deepsetnet}. The difficulty in such problems arises for two main reasons: because the \emph{cardinality} (i.e.\@ the number of elements) of the collection is unknown and can vary among different samples, and because collections when treated as a set are \emph{permutation-invariant}. However, canonical convolutional neural networks are not able to tackle these problems by design \cite{lee2019set}.

Transformers \cite{vaswani2017attention} have achieved impressive results on numerous collection prediction tasks \cite{carion2020end, lee2019set}, thanks to the attention module \cite{bahdanau2014neural}, a solution capable of aggregating information from the entire collection while also satisfying permutation invariance. 

The work in \cite{vaswani2017attention} first presents Transformers as an auto-regressive sequence-to-sequence model that generates a collection of output tokens one by one. Numerous following works \cite{chan2020imputer, ghazvininejad2019mask, gu2017non} introduce new variants of Transformers to reduce compute latency. However, the vast majority of auto-regressive Transformers focuses on sequential tasks only, such as machine translation \cite{ghazvininejad2019mask, gu2017non} or speech recognition \cite{chan2020imputer}. Recently, parallel Transformers have successfully been applied into computer vision, especially in object detection \cite{carion2020end, zhu2020deformable}, where the model outputs a set of bounding boxes \emph{in parallel}.  However, parallel Transformers rely on computationally intensive strategies, such as oversampling and Hungarian matching \cite{kuhn2005hungarian}.

The difference between auto-regressive and parallel Transformers is fundamental. Let us consider an image containing a collection of three objects (namely $A, B$ and $C$) which we are trying to detect. When asking a parallel Transformer to learn this task, we are actually asking the model to learn the joint probability of these three variables, conditioned on the model parameters $\Theta$, namely:
\begin{equation}\label{eq:joint_prob}
P(A, B, C \mid \Theta)
\end{equation}
In contrast, when the auto-regressive approach is presented with the same task, what it needs to learn is the chain of conditional probability distributions defined by:
\begin{equation}\label{eq:cond_prob}
P(A \mid \Theta) \cdot P(B \mid A, \Theta) \cdot P(C \mid A, B, \Theta)
\end{equation}

In the second case, the Transformer is asked to produce one element of the collection at a time, by exploiting information about the previous predictions. This is similar to machine translation, in which the Transformer outputs one word, conditioned on the previously generated output. \eq{joint_prob} and \eq{cond_prob} are equal by definition of the general product rule of probability, and Transformers are capable of modelling both. However, using \eq{cond_prob} with Transformers imposes an order even when there might not be a natural order present.


In this work, we study the difference between auto-regressive and parallel Transformers on polygonal shape prediction viewed as a collection of points, including individual points, lines, gates and polygons, as shown in \fig{datasets_examples}. Our contributions are: 
$(1)$ We test auto-regressive and parallel models  on four collection prediction datasets and one sequential dataset to provide the reader with a full picture of the advantages and disadvantages of both approaches.
$(2)$ We show that the conditional decomposition of the collection can be beneficial for Transformers when there is an explicit order in the elements of a set, such as predicting a collection of points on a line.
$(3)$ we show empirically that the conditional decomposition benefits from a particular order than others on a polygon dataset.

%% file: sections/1_related_work.tex
\section{Related work}


\textbf{Polygonal shape detection:} The work in \cite{girard2018end} views satellite image mapping as a polygon prediction task and makes use of a convolutional network to output the polygon vertices directly. 
The work in \cite{li2019topological} first detects bounding boxes of buildings, and then uses recurrent networks to extract vectorized building footprints. The work in \cite{girard2020polygonal} proposes to detect frame field for constructing building topology via fully convolutional networks and is able to precisely segment aerial images. Different from these works, we study the effect of self-attentions or Transformers in polygonal shape prediction.

\textbf{Object detection as set prediction.}
DETR \cite{carion2020end} first presents object detection as a set prediction task, and removes the need for non-maximal suppression on a large amounts of \emph{anchor boxes}. Similar to DETR, we also consider polygonal shapes in an image as a set. Different from DETR which only output bounding boxes, we produce vectorized polygons as a collection of vertices.



\textbf{Transformers.} Transformers were originally introduced by \cite{vaswani2017attention} as a novel auto-regressive, sequence-to-sequence model, and gained popularity thanks to their ability to dispense entirely with recurrence and support parallel processing of sequences. 
Their stunning results on machine translation and other language tasks \cite{brown2020language, devlin2018bert, liu2019roberta, radford2018improving} have recently shed a light on employing Transformers for computer vision tasks, such as image recognition \cite{dosovitskiy2020image}, object detection \cite{carion2020end}, segmentation \cite{ye2019cross}, set prediction \cite{lee2019set, wagstaff2019limitations} and other visual tasks \cite{chen2020pre, girdhar2019video, sun2019videobert, yang2020learning}. Inspired by these works, we study both auto-regressive and parallel Transformers on polygonal shape prediction.

\bigskip

\textbf{Transformers for set prediction.}
Deep Sets \cite{zaheer2017deep} has proven that transforming all elements of an input set into some latent representations and then combining them through a permutation invariant function is a universal approximator of any set function. The work in \cite{wagstaff2019limitations} shows that the attention layer of Transformers can also be viewed as a generalization of the sum operation of Deep Sets and is therefore also a universal approximator of set functions. Moreover, Lee \etal \cite{lee2019set} proposes an attention-based permutation-invariant framework which demonstrates superior results on set-structured data. Inspired by these works, we also study Transformers for set prediction task, but on a particular type of set data - polygonal shapes.



%% file: figures/model.tex
\begin{figure*}[t]
\begin{center}
\includegraphics[width=0.99\linewidth]{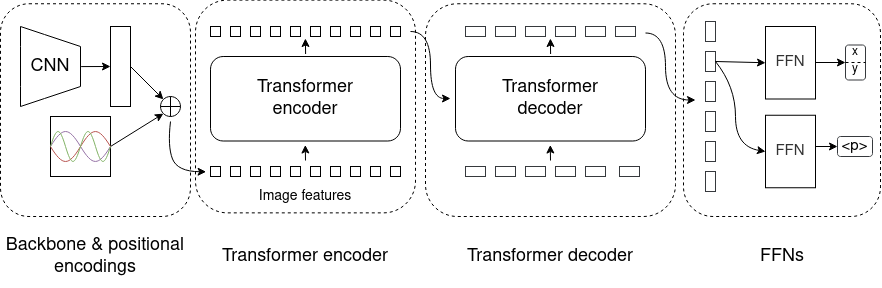}
\end{center}
  \caption{\textbf{The auto-regressive model for predicting polygons.} This model exploits the \emph{sentences of tokens} idea \cite{vaswani2017attention} which generates tokens auto-regressively. There are only four possible tokens in its vocabulary: \emph{start}, \emph{point}, \emph{end-of-polygon} and \emph{end}. The feed-forward network (FFN) at the end of the pipeline produces a class label and a pair of coordinates (x, y).  }
  
  
\label{fig:model}
\end{figure*}

%% file: figures/embeddings.tex
\begin{figure}[t]
\begin{center}
\includegraphics[width=0.99\linewidth]{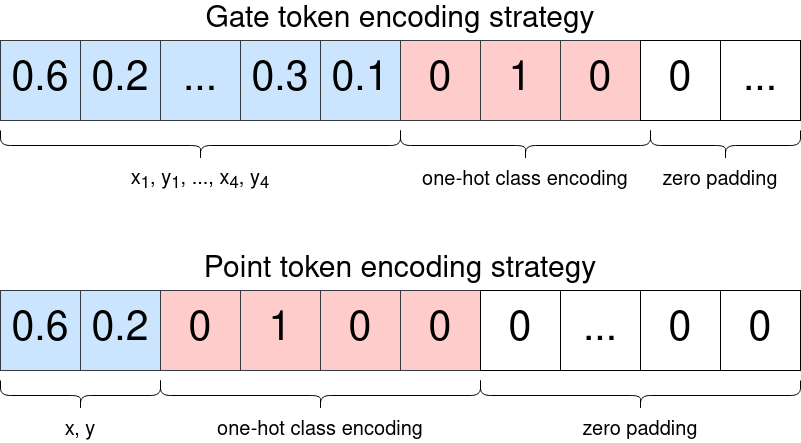}
\end{center}
   \caption{\textbf{Token embeddings for predicting gates and polygons}. The first $2n$ elements (in blue) of the vector represent the $(x,\,y)$ coordinates of the $n$ vertices/points, normalized by the image width and height. The following elements (in red) represent the class label in one-hot format. For gate detection, we have three red blocks because we have two special tokens \emph{start} and \emph{end}, and an object token \emph{gate}. For polygon detection we have 4 red blocks because all possible tokens are \emph{start}, \emph{end}, \emph{point} and \emph{end-of-polygon}, as explained in \ref{subsubsection:auto_reg}. The vector is padded with zeros to reach the required dimensionality of 256.}
\label{fig:embeddings}
\end{figure}


%% file: sections/2_method.tex
\section{Datasets}
To illustrate our setting, we first show examples of our datasets in \fig{datasets_examples}. We use 4 toy datasets and 1 synthetic dataset. The toy images are manually generated during training and if not specified otherwise, we generate 3 million images for training and 10,000 images for testing.

The \textbf{point} dataset contains images with $n$ white points randomly distributed over the image space. Each point's size is uniformly sampled from three possible values. The task is to predict the $x,\,y$ coordinates of all points in any order. This dataset is an instance of a pure set prediction problem, with points representing the set elements.

The \textbf{line} dataset contains images picturing a single white line composed by 7 segments. A green point of fixed size is placed on top of each end of a segment. The line is generated by first drawing a straight line going from the bottom left corner of the image to the top right corner, and then randomly shifting 8 equally distributed points perpendicularly to the line direction by $r \in [-15\%, 15\%]$ with respect to the image size. The task is to predict the $x,\,y$ coordinates of the 8 points following the line order, starting from the end on the bottom left of the image. The set elements are represented in this case by the green points, and this dataset is an example of a set prediction task where there exists an explicit order in which we need to predict its elements.

The \textbf{gate} dataset contains images with $n$ convex polygons of 4 corners, called \emph{gates}. Each gate is generated by defining 4 equally spaced points on a circumference of random radius $r \in [5\%, 40\%]$ with respect to the image size. Each point is then shifted randomly in the direction of the radius and the four points are finally connected to define the gate.
All four edges of the gates are of the same thickness, uniformly sampled out of 3 possible choices.

The \textbf{synthetic gate} dataset \cite{durnay2018detecting} contains realistic synthetic images generated with a graphical engine, which simulate the flight of a drone in different environments containing empty wireframe objects (EWFOs or \emph{gates}). This dataset is a simulation of the images that a UAV would face in the IROS 2018 Autonomous Drone Race \cite{jung2018direct}. The training set contains 26,000 images from two different scenes, with light coming from outside and inside the rooms. The test set contains 3,000 images from two additional scenes with light coming only from artificial sources placed inside the rooms. For all scenes, different walls and pavement textures are used, as well as different artificial shapes and light intensities for internal lamps. All images contain 1 to 4 gates. For both synthetic and toy gate datasets, the task is to predict the position of the four corners in the image for each gate. These two datasets represent a set prediction scenario closer to real-life problems, where the complexity of the single set element prediction (the gate) is greater than a simple point prediction. The synthetic gates dataset increases the complexity even further by picturing gates with different backgrounds and different lighting situations.

Finally, the \textbf{polygon} dataset contains images with $n$ polygons of $m \in [3, 7]$ corners, generated using the same technique as the gate dataset. As multiple non-convex polygons can be represented by the same set of points, the task is to predict the $m$ corners of a polygon in clockwise order. Any starting point is accepted, and distinct polygons can be predicted in any order. This dataset represents the hardest set prediction task in which we require our models to predict a \emph{set of ordered sets}.

For all datasets, $n$ is a parameter varied for different experimental settings. All samples in the toy datasets are RGB images of size 256x256, while images from the synthetic gate dataset are of size 400x400. The coordinates of the labels are represented as the relative height/width to the image size with values in $ [0, 1]$.

\section{Models}
All of the models that we implemented and tested are derived from DETR \cite{carion2020end}, an end-to-end Transformer which achieves competitive results against Faster R-CNN \cite{ren2015faster} on object detection. It takes advantage of a CNN backbone and parallel encoding-decoding Transformers to solve object detection as a set prediction task. In short, DETR is composed of four modules: a CNN backbone, a Transformer Encoder, a Transformer Decoder, and a feed-forward network (FFN). An example of our models is shown in \fig{model}. It pictures the auto-regressive variant used for the polygon toy setting dataset.

DETR takes as input an RGB image and extracts a high-level feature map of shape $C \times H\times W$, where $C$ is the size of number of channels, and $H$ and $W$ are the spatial dimensions. The feature map is supplemented with fixed positional embeddings before the Transformer encoder. The Transformer encoder is a stack of 6 self-attention mechanisms, each of which consists of a standard self-attention layer followed by a feed forward network, a residual connection and layer normalization \cite{carion2020end}. All of the models we tested are identical up to this stage of the architecture, but differ in the subsequent modules.

\subsection{Parallel models}

The design of parallel models is identical to DETR \cite{carion2020end}. The $N$ learned \emph{object queries} are converted \emph{in parallel} into $N$ output embeddings of size 256 by the Transformer decoder, which are subsequently fed into a simple feed forward network (FFN) for classification and position regression. 


Depending on the task, we modify the dimensionality of the FFN for position regression accordingly. For example, on the point and the line datasets, the output is a vector of size 2 representing the $(x,\,y)$ coordinates of a point normalized by the image width and height, while the output on the gate dataset is an 8-element vector indicating the four vertices of a gate in clock-wise order. The class labels for all datasets are the same, namely the \emph{object} class and the \emph{no-object} class.

On the polygon dataset, the FFN is replaced by a simple multi-layer Elman RNN  \cite{sherstinsky2020fundamentals}, since the output dimension is also a variable. The RNN model takes as input the embeddings from the \emph{object} class and generates a sequence of points one by one. 
The RNN is possibly the minimal  modification we could make to the architecture to work with polygons and have the smallest impact on the model's properties and our experiments.





\clearpage

\subsection{Auto-regressive models}\label{subsubsection:auto_reg}
The auto-regressive models diverge from the parallel ones on all set prediction tasks, because they work with \emph{sentences of tokens} that are generated one by one, by conditioning the next prediction on the previous ones. In particular, each token is a vector of dimension 256 that represents an element of the set, or acts as a special representation. Special tokens can be:
\begin{itemize}
    \item \emph{start} or \textbf{S}: This token is at the beginning of all sentences and defines the starts of the computation. 
    \item \emph{end} or \textbf{E}: This token is at the end of all sentences and terminates the computation.
    \item \emph{end-of-polygon} or \textbf{EOP}: This token separates polygons from polygons inside the same image. It acts similarly to the \emph{period} token used in machine translation to separate words belonging to different sentences.

\end{itemize}
Depending on the task, object tokens can be:
\begin{itemize}
    \item \emph{point} or \textbf{P}: On the point dataset and the line dataset, each point is a 2-element vector representing the position of the $(x,\,y)$ coordinate normalized by the image width and height.

    
    \item \emph{gate} or \textbf{G}: On both toy and synthetic gate datasets, a gate is represented as a vector of size 8, which defines the normalized coordinates $(x,\,y)$ of all 4 vertices.
    
\end{itemize}
To batch sequences of different lengths together, we pad sentences with the \emph{end} token up to the same length. For example, given a batched sequences where the maximal length is 8,  we pad the shorter sequence with 4 visible points only with two extra $\textbf{E}$:
\[\textbf{S},\;\textbf{P},\;\textbf{P},\;\textbf{P},\;\textbf{P},\;\textbf{E},\;\textbf{E},\;\textbf{E}\]

We use the same Transformer decoder as the parallel models, but adopt the \emph{masking} technique for parallel training, which prevents each token from attending to subsequent positions \cite{vaswani2017attention}. The token embedding is a fixed length vector of size 256 as shown in figure \ref{fig:embeddings}.



The auto-regressive approach predicts all polygons in a certain order. If not specified otherwise, we always predict objects going from left to right in the image, splitting ties with top-to-bottom order. We sort polygons and gates accordingly by their centers, computed as the average of their vertices. Particularly, for the polygon dataset, we also impose an order on \textbf{points} such that the polygon is clock-wisely defined by the points, otherwise the same set of the \textbf{points} may result in several polygons of different shapes.



%% file: sections/3_experiments.tex
\begin{figure}[t]
\begin{center}
\includegraphics[width=\linewidth]{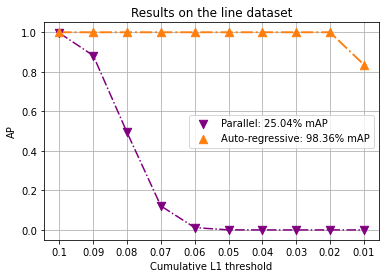}
\end{center}
   \caption{\textbf{Results on the line dataset.} The task is to predict a series of ordered points. The auto-regressive model produces approximately perfect prediction, while the parallel model fails.}
\label{fig:line_results}
\end{figure}

\section{Experiments}
First, we validate that the auto-regressive approach is preferable on the line dataset where there is an explicit natural order in the predictions, and that the parallel solution is better on the point dataset for pure set prediction tasks without orders. We then compare the two strategies on more challenging gate datasets where the element is no longer a single point and where data is scarce. Finally, we explore deeper into the auto-regressive solution to study upon the importance of the prediction order and the order of the conditional variables.  

On toy datasets, we train all models with 3,000,000 images. On the real gate datasets with 26,000 images, we train all models for 300 epochs. We apply multiple data augmentation techniques, including horizontal flip, vertical flip, hue shift, Gaussian noising.


We train all models on a single NVIDIA GeForce RTX 2080 Ti GPU with AdamW \cite{loshchilov2017decoupled}, and set the Transformer's learning rate to $10^{-4}$, the backbone's learning rate to $10^{-5}$, and the weight decay to $10^{-4}$. The learning rate is dropped by a factor of 10 after 200 epochs, or after 2,000,000 images for the toy settings. Mask R-CNN \cite{he2017mask} is also trained for 300 epochs with learning rate $5\times10^{-3}$, which is decreased by 10 after 200 epochs.


\subsection{Evaluation}

The evaluation metric on the gate and polygon datasets is mean Average Precision \cite{carion2020end}, averaged over different IoU thresholds ($[0.50, 0.55, \ldots, 0.95]$). On the point and line datasets, we also evaluate mean Average Precision, but averaged over different point-to-point $L_1$ distance thresholds ($[0.10, 0.09, \ldots, 0.01]$), computed directly on the relative coordinates.

\begin{figure}[t]
\begin{center}
\includegraphics[width=\linewidth]{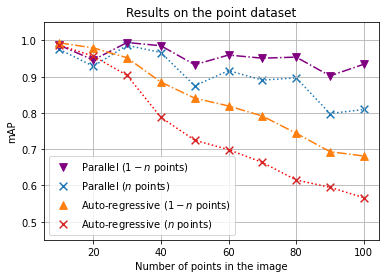}
\end{center}
   \caption{\textbf{Results on the point dataset.} The auto-regressive model shows marginally better performance than the parallel one on collections of limited cardinality, but its performance is lacking when the cardinality grows. The legends with \emph{$n$ points} represent results on images with exactly $n$ points while the others indicate that the number of points in an image varies from 1 to $n$.}

\label{fig:points_results}
\end{figure}

\subsection{Line and point detection}
With this experiment we show that the conditional decomposition of collections by auto-regressive Transformers is beneficial when the elements in a collection adhere to a natural order. 
Results on the line dataset are shown in figure \fig{line_results}. For this task, a line prediction is considered as a false positive if the sum of all $L_1$ point distances is greater than the given threshold. The experiment shows that in this setting the auto-regressive solution is much more precise than the parallel counterpart as it is able to achieve perfect mAP up to a threshold of 0.02.

On the other hand, the experiments on the point dataset study the behaviour of parallel and auto-regressive models on a pure set prediction task. The models are now expected to predict the $n$ points in any order. The parallel Transformers is order-insensitive because of the Hungarian matching, but the auto-regressive Transformers are trained by always feeding points in the left-to-right order. Results on this dataset are shown in \fig{points_results}. In this experiment, we train all models on images with 1 to $n$ points. We present test results on images with 1 to $n$ points, as well as results on images with exactly $n$ points. Results show that on collection prediction tasks, the auto-regressive approach is effective when the cardinality is low, but quickly deteriorates as the cardinality increases.

The two experiments on lines and points prove that the presence of a natural order in the task is indeed an important discriminative factor towards the performance of the parallel and auto-regressive models in collection prediction. Moreover, we show the advantage of the auto-regressive approach in low-cardinality prediction tasks. The parallel ones show significant advantage in predicting high-cardinality collections but at the cost of using redundant object queries (100 object queries in this experiment). 

\subsection{Gate detection}
\begin{figure}[t]
\begin{center}
\includegraphics[width=\linewidth]{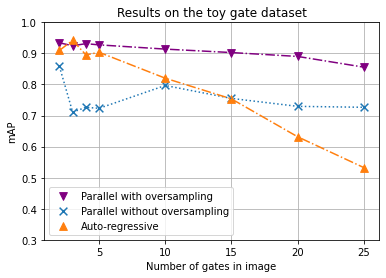}
\end{center}
   \caption{\textbf{Results on the gate dataset.} Each point of the plot is an experiment in which the model is trained and tested on images with 1 to $n$ gates. They show the necessity of oversampling for parallel Transformers. The performance of auto-regressive models deteriorates when collection cardinality increases.}
\label{fig:gates_ts_results}
\end{figure}

The following experiments verify whether the observations on the line and point datasets generalize in complicated scenarios where the collection element is a gate of four vertices. First, we evaluate the performances of auto-regressive approach and its parallel counterpart on the toy gate dataset, as shown in \fig{gates_ts_results}. There are two parallel models in comparison: one with oversampling where we use 30 object queries which is several times more than the total number of gates, and the other one without oversampling where we use as many object queries as the maximal number of objects in the images. The parallel model with oversampling outperforms the one without oversampling substantially, validating the need for the
parallel models to oversample the cardinality.  Moreover, we show once more the auto-regressive approach performs comparably to the parallel one on low-cardinality collections, but suffers from the growing cardinality.

\begin{figure}[t]
\begin{center}
\includegraphics[width=\linewidth]{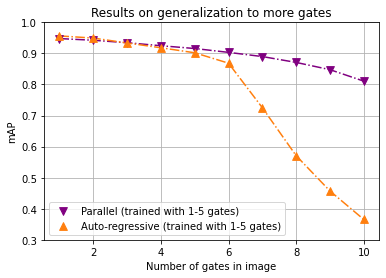}
\end{center}
   \caption{\textbf{Generalization capability of the different models.} We test both parallel and auto-regressive on images with twice as many gates as the training images. The parallel model is able to detect many more gates than the auto-regressive one. We believe oversampling attributes to better generalization.}
\label{fig:generalization}
\end{figure}

In \fig{generalization}, we provide our findings on the generalization capabilities of these models, where the number of gates in test images is up to twice more than the number during training. The auto-regressive model struggles at detecting more gates, while the parallel one only shows minor performance decrease and achieves an mAP of 0.8 even on images with twice the amount of gates. We believe this behavior is a result of the oversampling strategy: each of the 30 object queries is assigned to at least one gate during training, which implicitly tells the model that there could be more objects than the ones it sees in each image.

Finally, we numerically compare the general performances of all models on both toy and synthetic gate datasets with images containing 1 to 6 gates. We choose Mask R-CNN as the baseline. \tab{results_gates} shows parallel Transformer outperforms Mask R-CNN by 5 $\%$ mAP on the toy dataset and by 4 $\%$ mAP on the synthetic gate dataset. The auto-regressive approach is not able to achieve comparable results as there is no natural order in this task. However, the auto-regressive model is approximately two times faster than the parallel model as it does not require Hungarian algorithm or oversampling.


\begin{table*}[!htbp]
    \begin{center}
    \begin{tabular}{p{2.5cm}|cc|cc}
    \toprule
    \multicolumn{1}{c|}{Models} & Toy gates & Training time [min/10k images] & Synthetic gates & Training time [min/10k images] \\  
    \midrule
    \emph{Mask R-CNN} & 90.67\% & 23.6 &  61.30\% & 13.85 \\
    Parallel & \textbf{95.45\%} & 2.2 & \textbf{65.00\%} & 3.85 \\
    Auto-regressive & 87.67\% & \textbf{1.5} & 59.03\% & \textbf{1.85} \\
    \bottomrule
    \end{tabular}
    \end{center}
    \caption{\textbf{Numerical results on gate datasets}. We evaluate the overall performance in mAP averaged over 10 IoU thresholds [0.50, 0.55, \ldots, 0.95]. The parallel model outperforms Mask R-CNN. The performance of auto-regressive model is lacking due to the absence of natural order in this task.}
    \label{tab:results_gates}
\end{table*}

\subsection{Polygon detection}
Results on the polygon toy setting dataset are shown in table \tab{results_polygons}. This experiment explores an hybrid scenario in which the collection elements we are trying to predict are also collections with an imposed order, because a collection of points can represent multiple polygons if the order of its points is not given. The auto-regressive approach is the special one described at the end of section \ref{subsubsection:auto_reg}, which predicts a polygon as a sequence of points followed by the \emph{end-of-polygon} token, and it outperforms the parallel model substantially by over 20 $\%$ in terms of mAP. When using the auto-regressive solution on polygons, we are predicting each polygon individually by predicting its vertices and then an end-of-polygon token. This means that we are splitting a problem of directly predicting all points of all polygons into multiple sub-problems of predicting each polygon and then deciding when all points have been predicted. We speculate that the advantage of the auto-regressive model could be a direct consequence of the conditional decomposition of the joint probability, as imposing the conditional order on these models can serve as a strong inductive bias by reducing the search space of the model. This behaviour might not show up on the previous experiments because gates, points and lines are simple collection elements which do not enlarge the search space enough. As polygon detection is representative of many common computer vision tasks, we leave further exploration of this approach as future work.


\begin{table*}[!htbp]
\begin{center}
\begin{tabular}{p{2.5cm}cc}
\toprule
\multicolumn{1}{c}{Models} & Polygons dataset & Training time [min/10k images]\\
\midrule
Parallel & 53.55\% & 5.1\\
Auto-regressive & \textbf{76.41}\% & \textbf{2.15}\\
\bottomrule
\end{tabular}
\end{center}
\caption{\textbf{Numerical results on the polygon dataset.} Scores are represented as mAP averaged over 10 IoU thresholds [0.50, 0.55, \ldots, 0.95]. The auto-regressive approach outperforms the parallel one by over 20\% absolute mAP. We speculate that this could be a direct consequence of conditional decomposition of the collection.}
\label{tab:results_polygons}
\end{table*}

\subsection{Orders of conditional decomposition}

\begin{table*}[!htbp]
\begin{center}
\begin{tabular}{p{6.5cm}cc}
\toprule
 & With positional encodings & W/o positional encodings \\
\midrule
Order based on polygon positions & $-1.90\% \pm 1.07\%$ & -- \\
Order based on the size of polygons & $-7.68\% \pm 4.53\%$ & $-2.95\% \pm 1.30\%$ \\
\bottomrule
\end{tabular}
\end{center}
\caption{\textbf{Ablation studies on position encodings and condition orders.} This tables shows results of adding positional encodings and imposing orders. The top performing model is spatially ordered (left-to-right and top-to-bottom) and has no positional encoding. Our observations are that: (1) changing the imposed order on auto-regressive Transformers has a great impact on the performance as shown in the first column; (2) adding position encodings decreases the performance.}
\label{tab:results_order}
\end{table*}


We conclude our experiments by providing insights on the importance of orders imposed on conditional decomposition of a collection by the auto-regressive Transformers. Moreover, we also study the influence of positional endcodings. 
We run experiments on the toy gate dataset, synthetic gate and the polygon datasets multiple times, by using two different object orderings and by adding or removing the positional encodings of the attention layer. Without the positional encodings, the Transformer decoder is unaware of the index of the previous predictions in the sequence. When predicting a new point, it only knows which points were predicted before, but not their orders. The two orderings are the left-to-right, top-to-bottom order and the small-to-large order in terms of the size of the objects.
Experiment results are shown in table \tab{results_order}. Using the left-to-right, top-to-bottom order and removing the positional encodings lead to the best performance in all experiments. We show the mean absolute difference of other models compared to best one. The result shows that the artificial order \emph{matters} for auto-regressive models. Moreover, we show that fixed positional encodings are not beneficial in this setting, in contrast to the original Transformer architectures. This is expected as different orders of the same words in NLP can have different meaning, while this does not matter in our setting, as knowing the location of a point can already prevent our model to predict the same point twice. 

%% file: sections/4_conclusion.tex
\section{Conclusion}
We studied two important variants of the Transformer architecture: the parallel one and the auto-regressive one on the polygonal shape prediction task cast as a collection prediction problem. We find that the ordering of objects is a strong factor towards the performance of these models on point, line, gate and polygon datasets. The auto-regressive model benefits  collections of small cardinality or on collection prediction problems that can be easily split into multiple easier tasks.



One limitation of this research is that most experiments are conducted on toy datasets only, and it is unclear whether the observations and conclusions on toy datasets generalize on challenging real-world datasets. A task that would be suitable to expand our research is polygonal building segmentation from satellite images, as in \cite{girard2020polygonal, mohanty2020deep}.


As future work, we find it important to further explore on the polygon detection task as it is a fundamental problem that is representative in computer vision. Testing the effect of fixed or learned token embeddings would also be of interest.
